# Comment on "A note on Over-Smoothing for Graph Neural Networks"

Reuven Guetta          Razi Hasson

Technion - IIOT

## Abstract

The paper "A Note on Over-Smoothing for Graph Neural Networks" addresses a critical issue in the field of Graph Neural Networks (GNNs), namely, the phenomenon of over-smoothing. Over-smoothing occurs when the performance of GNNs deteriorates as the number of layers increases, resulting in embeddings that become indistinguishable from one another. Previous studies [Oono & Suzuki, 19] primarily focused on linear cases. however, this paper extends the analysis to more general GNN architectures. This paper demonstrates that under certain conditions related to the spectrum of the augmented normalized Laplacian, the Dirichlet energy of the node embeddings converges to zero, leading to a loss of discriminative power. The paper provides a clean and intuitive way to measure the expressiveness of embeddings through Dirichlet energy, offering simpler proofs (compared to earlier work) and handling non-linearities more effectively.

## 1. Introduction

The introduction of the paper outlines the development and challenges of Graph Neural Networks (GNNs), which are a class of neural networks implemented to graph-structured data. It highlights the success of various GNN models like Graph Convolutional Networks (GCNs) and their variants, such as GraphSAGE, GAT, and SGC, in semi-supervised classification. Despite their success, GNNs face significant challenge particularly with increasing depth. Stacking more layers leads to performance deterioration due to issues like vanishing gradients, overfitting, and a phenomenon called over-smoothing. Over-smoothing, first identified by [Li et al., 2018], refers to node features converging to similar values as more layers are added, reducing the model's ability to distinguish nodes.

The paper builds on the work of [Oono & Suzuki, 19], which analyzed over-smoothing in linear GNNs and extended it to non-linear cases using ReLU. The goal of this paper is to further this analysis by introducing a method to measure the expressiveness of node embeddings using Dirichlet energy.

Advantages of this approach: conceptually clean, recovers results from [Oono & Suzuki, 19], easily applies to Leaky ReLU, extendable to common nonlinearities for regular graphs, requires only elementary linear algebra.

The paper contributions include experiments that examine how edge operations (dropping edges or increasing weight) affect Dirichlet energy, offering insights into increasing GNN expressiveness.

## 2. Notation and GCN architecture modification resulting in spectral polynomial GNNs

Graph definitions: $A, D \in \mathbb{R}^{N \times N}$, $A$ is called the adjacency matrix, $D$ is called the degree matrix of graph G. define $\tilde{A} := A + I_N$, $\tilde{D} := D + I_N$ adjacency and degree matrices of graph G augmented with self-loops. $\tilde{\Delta} := I_N - \tilde{D}^{-\frac{1}{2}} \tilde{A} \tilde{D}^{-\frac{1}{2}}$ the augmented normalized Laplacian of G, set $P := I_N - \tilde{\Delta}$.

GCN architecture: let $L, C \in \mathbb{N}_+$ be the layer and channel sizes. $f = f_L \circ \ldots \circ f_1$, GCN function defined as a composition of layer functions. $f_l(X) := MLP_l(PX)$, $fl : R^{N \times C^l} \to R^{N \times C_{l+1}}$, the layer $l$ function. $MLPl(X) := \sigma(\cdots \sigma(\sigma(X)W_{l1})W_{l2} \cdots W_{lH_l})$, multi-layer perception with non-linear activation $\sigma$. (weight matrix $W_l$ isn't necessarily squared).

Embeddings: embedding at layer $l + 1$ is defined as by: $X^{(l+1)} := f_l(X^{(l)})$, $X^{(0)}$ is the initial embedding.

Our interest: the asymptotic behavior of the output $X^{(L)}$ as $L \to \infty$. Additionally, by a lemma we know that eigenvalues of $\tilde{\Delta} \in [0,2]$, eigenvalues of $P = I_N - \tilde{\Delta} \in (-1,1]$. which is important for understanding the convergence properties of node embeddings as layers are stacked.

## 3. Main result:

The authors present their key findings regarding the behavior of GCNs as the number of layers increases. The primary focus is on the Dirichlet energy, which measures the "smoothness" of node embeddings and is directly tied to over-smoothing in GNNs. Their main contribution is showing that the Dirichlet energy decreases exponentially with respect to the number of layers (later, Theorem 3.4), resulting in embeddings that lose discriminative power.

The Dirichlet energy $E(f)$ of a scalar function $f \in \mathbb{R}^{N \times 1}$ on graph G is defined by: $E(f) = f^T \tilde{\Delta} f$, and for vector field $X_{N \times c} = [x_1, \ldots, x_N]^T$, where $x_i \in \mathbb{R}^{1 \times c}$ we define Dirichlet energy by $E(f) = tr(X^T \tilde{\Delta} X)$. Each layer of the GCN is represented by $\mathbf{f_l}(X) = \sigma(\cdots \sigma(\sigma(PX)W_{l1})W_{l2} \cdots W_{lH_l})$.

By 3 key lemmas, we examine how different components of GCNs $(P, W_l, \sigma)$ affect the Dirichlet energy:

**Lemma 3.1.** $E(PX) \leq (1 - \lambda)^2 E(X)$ where $\lambda$ is the minimal non zero eigenvalue of $\tilde{\Delta}$.

**Lemma 3.2.** $E(XW) \leq \|W^T\|_2^2 E(X)$.

Note: The Proof of this lemma does not rely on the specific structure of the graph, as it only leverages the fact that the Laplacian is positive semi-definite matrix. Which makes the result more general. In addition, the proof can be extended to weight matrices of varying dimensions, whereas the paper [Oono & Suzuki ,2019] assumes that the embedding dimension is fixed across layers.

**Lemma 3.3.** $E(\sigma(X)) \leq E(X)$, for ReLU or leaky-ReLU.

In other words: ReLU or Leaky-ReLU either preserve or decrease the Dirichlet energy.

Note: For regular graphs, the lemma can be extended to include other non-linearities such as Tanh and Sigmoid. Also note that the proof of the lemma (as presented in the paper) relies on specific property of ReLU and Leaky-ReLU, thus, for activations Tanh and Sigmoid the above inequality doesn't necessarily hold in general.

Combining previous lemmas, we get:

**Theorem 3.4.** denote by $s_{lh}$ the maximum singular value of $W_{lh}^T$, set $s_l := \prod_{h=1}^{H_l} s_{lh}$, $\bar{\lambda} := (1-\lambda)^2$. Then, for any $l \in \mathbb{N}_+$ we have: $E(\mathbf{f}_l(X)) \leq s_l \bar{\lambda} E(X)$.

**Corollary 3.5.** denote $s := \sup_{l \in \mathbb{N}_+} s_l$, we have: $E(X^{(L)}) \leq O\left((s\bar{\lambda})^l\right)$, in particular, $E(X^{(L)})$ converges to 0 when $s\bar{\lambda} < 1$.

This result directly supports the over-smoothing phenomenon, where deeper GCNs becomes less expressive. The paper generalizes the previous work in [Oono & Suzuki ,2019], extending it to cover a broader set of non-linearities and graph structures.

## 4. Key Differences

The differences between this paper and [Oono & Suzuki ,2019] can be examined from 2 key perspectives:

- Handling varying dimensions:
  [Oono & Suzuki ,2019] focused on cases where the embedding dimension remains constant across layers. They introduced a measure $d_\mathcal{M}(X)$ which captures the distance between node embedding and a subspace $\mathcal{M}$ of $\mathbb{R}^{N \times C}$, $\mathcal{M} := U \otimes \mathbb{R}^C$ where $U$ is null space of a normalized graph Laplacian (see [Oono & Suzuki ,2019]). Whereas this paper modifies the definition of $\mathcal{M}$ to handle varying embedding dimensions across layers. We define $\mathcal{M} = U \otimes \mathbb{R}^C$, $\mathcal{M}' = U \otimes \mathbb{R}^{C'}$ where $U$ is as previous and $c' \neq c$. With this modification, the inequality $d_{\mathcal{M}'}(XW) \leq s d_\mathcal{M}(X)$ now holds (where $s$ is a singular value of $W \in \mathbb{R}^{c \times c'}$). This modification makes the analysis more applicable to GNNs, which often have changing dimensions between layers.

- Activation functions non-linearity:

[Oono & Suzuki ,2019] restricted their analysis to GNNs with ReLU activation. Their results rely on ReLU's mathematical property, that it is a projection onto the cone $\{X > 0\}$, thus aligning well with the non-negativity of the eigenvectors belonging to the largest eigenvalues. This reliance made it complicated to extend their result to activations as Leaky ReLU or Sigmoid.

In contrast, this paper provides a simpler, more general analysis. And the proof of Lemma 3.3 becomes straightforward, as Dirichlet energy can be broken down into terms, where each term can be easily analyzed for various activation functions.

## 5. Experiments

The author presents several experiments done to check the effects of basic edge operations, removing nodes and increasing weights, on the Dirichlet energy. In the first couple of experiments, they checked it by computing the eigenvalues before and after dropping and adding weights randomly to nodes, and by removing and adding nodes, also at random, and calculating the Dirichlet energy of the new graph.

The authors make the following observations:

1- For most graphs and ratios (except for certain cases in the Cora and CiteSeer datasets), removing edges tends to increase the Dirichlet energy of the graph, and thus helps relive over-smoothing.
2- In most situations, increasing an edge's weight (from 1 to 10,000) and dropping edges seems to have a 'dual' relationship. In other words, raising the weight of a few edges to very high values has a similar effect to removing a lot of edges when considering eigenvalues and Dirichlet energy. Thus, the authors hypothesize that increasing the weight of a few edges to a high value will help relieve over-smoothing.

## 6. Conclusion

The paper presents an alternative simple proof that graph neural networks (GNNs) lose expressive power exponentially, similar to the findings of [Oono & Suzuki ,2019], but also extending the analysis to Leaky ReLU activations.

The authors identify several key challenges and future research directions:

The key challenge of analyzing the over-smoothing effect lies in the non-linearity. One way of tackling this problem is in the article, but perhaps it could be extended to more general graph learning. They also note the importance of the assumption on the norm of the weight function in GNNs, which is crucial (though possibly too strict) in their proof. They also suggest that since Dirichlet energy is well studied in the context of graph sparsification, some results can be used in the context of GNNs. Though this paper helps tackle over smoothing, the real over-smoothing effect, measured by the Rayleigh quotient for deep GNNs, remains an open and significant problem to explore.

# 7. Modification of GNN and a new proposition and proof of lemma 3.3 for Leaky-ReLU

Modification:

> To modify the GCN architecture as requested, we replace the multiplication by P with a general polynomial of $\tilde{\Delta}$ for each layer $l$, as follows:
> Original GCN layer: $f_l(X) := MLP_l(PX)$, where $P := I_N - \tilde{\Delta}$.
> Modified GCN layer: $f_l(X) := MLP_l(P_l(\tilde{\Delta})X)$, where $P_l$ is a polynomial of $\tilde{\Delta}$ specific to layer $l$, defined as:
> $$P_l(\tilde{\Delta}) = a_{0l}I_N + a_{1l}\tilde{\Delta} + \cdots + a_{kl}\tilde{\Delta}^k$$
> the coefficients $a_{0l}, \ldots, a_{kl}$ are specific to layer $l$.
> The GCN structure remains as defined previously, except now, each $f_l$ uses its own polynomial $P_l(\tilde{\Delta})$.

## Proposition 7.1 (Razi-Reuven 2024)

*Let $\{\lambda_i\}$ be the eigenvalues of $\tilde{\Delta}$, and denote by $\lambda$ the minimal non-zero eigenvalue of $\tilde{\Delta}$. suppose that there exists a layer $L$ and $\epsilon \in (0,1)$ such that $\forall l > L$, $P_l(x)$ is monotonically decreasing and $\forall \lambda_i$ we have $s_l P_l(\lambda_i)^2 < 1 - \epsilon$. Then,*

*$E(f_l X) \leq s_l P_l(\lambda)^2 E(X)$. In particular, $E(X^l) \leq O((sP_l(x)^2)^l) < O((1-\epsilon)^l)$ and converges exponentially to 0.*

*Proof:* our proof relies on the previous lemmas with a bit modification to Lemma 3.1. We start by proving the following version of Lemma 3.1:

**Lemma 7.2.** Let $\lambda$ be the minimal non-zero eigenvalue of $\tilde{\Delta}$, and $l \geq L$, then $E(P_l(\tilde{\Delta})X) \leq P_l(\lambda)^2 E(X)$.

*Proof:* Let us denote the eigenvalues of $\tilde{\Delta}$ by $\lambda_1 \ldots \lambda_n$ and $v_1, \ldots, v_n$ the associated eigenvectors of length 1. Define $f = \sum c_i v_i$, where $c_i \in \mathbb{R}$.

First, we know that a power and a sum of symmetric matrices is a symmetric matrix. Thus, since $\tilde{\Delta}$ is symmetric by definition, $P_l(\tilde{\Delta})$ is also a symmetric matrix. Therefore, we have:

$$E(P_l(\tilde{\Delta})f) = f^T P_l(\tilde{\Delta})^T \tilde{\Delta} P_l(\tilde{\Delta}) f = f^T P_l(\tilde{\Delta}) \tilde{\Delta} P_l(\tilde{\Delta}) f =$$
$$= \sum c_i^2 \lambda_i P_l(\lambda_i)^2 \leq P_l(\lambda)^2 E(f)$$

Where the last inequality holds since $P$ is monotonously decreasing and since $\lambda$ is the minimal non-zero eigenvalue. Thus we get $E(PX) \leq P_l(\lambda)^2 E(X)$. ∎

Now, since Lemma 3.2 and Lemma 3.3 hold regardless of P, we have $E(XW) \leq \|W^T\|_2^2 E(X)$, and $E(\sigma(X)) \leq E(X)$, for ReLU or leaky-ReLU.

By assuming that for all $l \geq L$ we have that $\forall \lambda_i : s_l P_l(\lambda_i)^2 < 1 - \epsilon$, similarly to Theorem 3.4 we get that $E(\mathbf{f}_l(X)) \leq s_l P_l(\lambda)^2 E(X)$.

Thus from Corollary 3.5 we get, $\forall l \geq L : E(X^{(l)}) \leq O((sP_l(\lambda)^2)^l) < O((1-\epsilon)^l)$ and converges exponentially to 0 since $s_l P_l(\lambda)^2 \leq sP_l(\lambda)^2 < 1 - \epsilon$. Which completes our proof. ∎

**Proof of Lemma 3.3 for Leaky-ReLU:**

Definition of Leaky-ReLU:

$$LeakyReLU(x) = \begin{cases} x, & x > 0 \\ ax, & x \leq 0 \end{cases}, where\ a \in (0,1).$$

**Lemma 3.3.** $E(\sigma(X)) \leq E(X)$, for Leaky-ReLU.

*Proof.* Denote $LeakyReLU(x) := \sigma(x)$, and note that $\sigma'(x) = \begin{cases} 1, & x > 0 \\ a, & x \leq 0 \end{cases}$, thus the Lipshitz constant is $\max(1, |a|) = 1$. In addition, $\sigma(cx) = c\sigma(x)\ \forall c \in \mathbb{R}^+\ and\ x \in \mathbb{R}$. Thus we get:

$$|c_1 a - c_2 b| \geq |\sigma(c_1 a) - \sigma(c_2 b)| = |c_1 \sigma(a) - c_2 \sigma(b)|$$

Thus, since $E(f) = \sum_{(i,j) \in E} w_{i,j} \left( \frac{f_i}{\sqrt{1+d_i}} - \frac{f_j}{\sqrt{1+d_j}} \right)^2$, and by replacing $c_1, c_2, a, b$ with $\frac{1}{\sqrt{1+d_i}}, \frac{1}{\sqrt{1+d_j}}, f_i, f_j$ we get that it holds for scalar field $f$, $E(\sigma(f)) \leq E(f)$.

By extending this to vector field $X$, we complete the proof. ∎

## 8. Concluding Remarks:

The paper offers a blend of theoretical and practical contributions to tackling over-smoothing in Graph Neural Networks (GNNs), which has been a challenging issue in the field. We found the authors use of Dirichlet energy to explain over-smoothing both intuitive and impactful. By framing over-smoothing in terms of Dirichlet energy, the authors have given us a quantitative lens to look at over-smoothing, which we think is invaluable. This approach not only deepens our understanding but also opens new possibilities for how we might mitigate this problem with a more precise approach.

Another aspect we appreciate is their focus on non-linear activation functions, By expanding their analysis to nonlinearities, particularly ReLU and Leaky ReLU, this paper becomes highly relevant for practical applications since most real-world GNN implementations use such activation functions.

The practical suggestions in the paper, specifically the manipulation of edge weights and configurations are also a highlight for us. This is where the paper goes beyond a purely theoretical contribution and provides a way to experiment and see immediate results.

# Contact Info

Razihasson@campus.technion.ac.il